\documentclass[11pt,a4paper]{article}
\usepackage[hyperref]{acl2017}
\usepackage{times}
\usepackage{latexsym}

\usepackage{url}

\aclfinalcopy 

\usepackage{amssymb}
\usepackage{amsthm}
\usepackage{color}
\usepackage{amsmath}
\usepackage{graphicx}
\usepackage{multirow}
\usepackage{breqn}
\usepackage{url}
\usepackage{comment}

\usepackage{subcaption}
\usepackage{caption}
\usepackage{tikz}
\usepackage{verbatim}

\usepackage{mathrsfs}
\usepackage{verbatim}
\usepackage{times}
\usepackage{url}
\usepackage{latexsym}
\usepackage{color}
\usepackage{float}
\usepackage{array}
\usepackage{amsmath,amsfonts,amssymb}
\usepackage{algorithm}
\usepackage[noend]{algpseudocode}
\usepackage{graphicx}
\usepackage{subcaption}
\usepackage{booktabs}
\usepackage{multirow}
\usepackage{tikz,pgfplots}
\usepackage{pgfplotstable}
\usepackage{enumitem}
\usepackage{pifont}
\usepackage{fancyvrb}
\usepackage{footnote}
\makesavenoteenv{tabular}
\makesavenoteenv{table}

\usetikzlibrary{arrows,shapes,positioning,shadows,trees}

\usepackage[font=small]{caption}

\renewcommand{\cite}[1]{\citep{#1}}

\usepackage{stmaryrd}

\def\by{{\boldsymbol{y}}}
\def\bx{{\boldsymbol{x}}}

\newcommand{\mnote}[1]
{
}

\pgfplotscreateplotcyclelist{mark list*}{%
every mark/.append style={solid,fill=.!80!black},mark=*\\%
every mark/.append style={solid,fill=.!80!black},mark=square*\\%
every mark/.append style={solid,fill=.!80!black},mark=triangle*\\%
every mark/.append style={solid,fill=.!80!black},mark=halfsquare*\\%
every mark/.append style={solid,fill=.!80!black},mark=pentagon*\\%
every mark/.append style={solid,fill=.!80!black},mark=halfcircle*\\%
every mark/.append style={solid,fill=.!80!black,rotate=180},mark=halfdiamond*\\%
every mark/.append style={solid,fill=.!80!black!40},mark=otimes*\\%
every mark/.append style={solid,fill=.!80!black},mark=diamond*\\%
every mark/.append style={solid,fill=.!80!black},mark=halfsquare right*\\%
every mark/.append style={solid,fill=.!80!black},mark=halfsquare left*\\%
}
\pgfplotscreateplotcyclelist{color list}{%
red,blue,black,yellow,brown,teal,orange,violet,cyan,green!70!black,magenta,gray}
\pgfplotscreateplotcyclelist{linestyles}{solid,dashed,dotted}
\pgfplotscreateplotcyclelist{linestyles*}{solid,dashed,dotted,dashdotted,dashdotdotted}

\tikzset{every mark/.append style={scale=1.5}}

\pgfplotscreateplotcyclelist{mylist}{%
{blue,mark=*,solid},
{green!60!black,mark=triangle},
{brown!60!black,mark=diamond*},
{brown!60!black,mark options={fill=brown!40},mark=otimes*},
{yellow!60!black,mark=triangle*,mark options={fill=brown!40}},
{red,mark=square,}}

\newcommand*{\Let}[2]{\State {#1} $\gets$ {#2}}
\newcommand\Gap{~~~~~~~}

\title{Beyond Bilingual: Multi-sense Word Embeddings using Multilingual Context}

\author{Shyam Upadhyay$^{1}$\Gap Kai-Wei Chang$^2$\Gap Matt Taddy$^3$\Gap Adam Kalai$^3$\Gap James Zou$^4$\\
    $^1$University of Illinois at Urbana-Champaign, Urbana, IL, USA \\
      $^2$University of Virginia, Charlottesville, VA, USA\\
      $^3$Microsoft Research, Cambridge, MA, USA\\
      $^4$Stanford University, Stanford, CA, USA\\
      {\tt upadhya3@illinois.edu}, {\tt kw@kwchang.net} \\
      {\tt \{taddy,adum\}@microsoft.com}, {\tt jamesyzou@gmail.com}
}

\date{}

\begin{document}
\maketitle
\begin{abstract}
Word embeddings, which represent a word as a point in a vector space, have
become ubiquitous to several NLP tasks. 
A recent line of work uses bilingual (two languages) corpora to learn a different vector for each sense of a word, by exploiting crosslingual signals to aid sense identification.
We present a multi-view Bayesian non-parametric algorithm which improves multi-sense word embeddings by 
(a) using multilingual (i.e., more than two languages) corpora to significantly improve sense embeddings beyond what one achieves with bilingual information, and (b) uses a principled approach to learn a variable number of senses per word, in a data-driven manner.
Ours is the first approach with the ability to leverage multilingual corpora efficiently for multi-sense representation learning.
Experiments show that multilingual training significantly improves performance over monolingual and bilingual training, by allowing us to combine different parallel corpora to leverage multilingual context. Multilingual training yields comparable performance to a state of the art monolingual model trained on five times more training data.


\end{abstract}

\section{Introduction}
\label{sec:introduction}
Word embeddings~\citep[{\em inter alia}]{turian:2010,mikolov-yih-zweig:2013:NAACL} represent a word as
a point in a vector space. This space is able to capture semantic relationships: vectors of words with similar meanings have high
cosine similarity~\cite{Turney:2006:SSR:1174520.1174523,turian:2010}.
Use of embeddings as features has been shown to benefit several NLP tasks
and serve as good initializations for deep architectures ranging from dependency parsing \cite{bansal:2014} to named entity
recognition \cite{guo2014}.

Although these representations are
now ubiquitous in NLP, most algorithms for learning word-embeddings do not allow a word to have different meanings in
different contexts, a phenomenon known as polysemy. For
example, the word {\em bank} assumes different meanings in financial
(eg. ``bank pays interest'') and geographical contexts (eg. ``river
bank'') and which cannot be represented adequately with a single
embedding vector.  Unfortunately, there are no large sense-tagged corpora available 
and such polysemy must be inferred from the data during the embedding process.

\begin{figure}[h]
  \centering
  \includegraphics[scale=0.32]{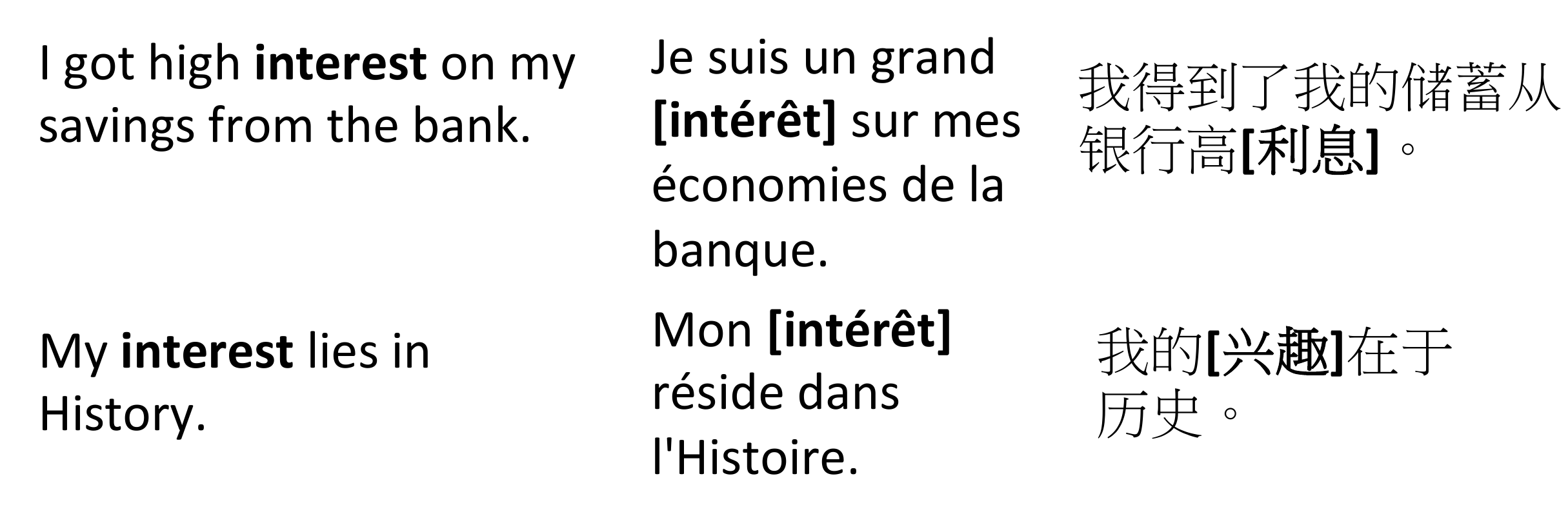}
  \caption{{\bf Benefit of Multilingual Information (beyond bilingual)}: Two different senses of the word ``interest'' and their translations to French and Chinese (word translation shown in {\bf [bold]}). While the surface form of both senses are same in French, they are different in Chinese.}
    \label{fig:multiling}
\end{figure}

Several
attempts~\cite{reisinger:naaclx10,neelakantan-EtAl:2014:EMNLP2014,li2015hierarchical}
have been made to infer multi-sense word representations by modeling
the sense as a latent variable in a Bayesian non-parametric
framework. These approaches rely on the "one-sense per collocation"
heuristic~\cite{yarowsky1995unsupervised}, which assumes that presence
of nearby words correlate with the sense of the word of interest. This
heuristic provides only a weak signal for sense identification, and such
algorithms require large amount of training data to achieve
competitive performance.


Recently, several approaches~\cite{guo2014learning,vsuster-titov-vannoord:2016:N16-1} propose to learn
multi-sense embeddings by exploiting the fact that different senses of
the same word may be translated into different words in a foreign
language~\cite{dagan1994word,resnik1999distinguishing,diab-resnik:2002:ACL,ng2003exploiting}. For
example, {\em bank} in English may be translated to {\em banc} or
{\em banque} in French, depending on whether the sense is financial or
geographical.  Such bilingual distributional information allows the
model to identify which sense of a word is being used during training.
 
However, bilingual distributional signals often do not suffice. It is common that polysemy for a
word survives translation.
Fig. \ref{fig:multiling} shows an illustrative example -- both
senses of {\em interest} get translated to {\em int\'er\^et} in French. 
However, this becomes much less likely as
the number of languages under consideration grows. By looking at Chinese translation in Fig. \ref{fig:multiling}, we can observe that the
senses translate to different surface forms. Note that the opposite
can also happen (i.e. same surface forms in Chinese, but
different in French).
Existing crosslingual approaches are inherently bilingual and cannot naturally extend to include
additional languages due to several limitations (details in Section~\ref{sec:multi-ext}).
Furthermore, works like ~\cite{vsuster-titov-vannoord:2016:N16-1} sets a fixed number of senses for each word, leading to inefficient use of parameters, 
and unnecessary model complexity.\footnote{Most words in conventional English are monosemous, i.e. single sense (eg. the word {\em monosemous})}

This paper addresses these limitations  by proposing a multi-view
Bayesian non-parametric word representation learning algorithm which
leverages multilingual distributional information.
Our representation learning framework is the first
multilingual (not bilingual) approach, allowing us to
utilize arbitrarily many languages to disambiguate words in English.  
To move to multilingual system, it is necessary to ensure that the embeddings of each foreign language are relatable to each other (i.e., they live in the same space). We solve this by proposing an algorithm in which word representations  are learned {\it jointly} across languages, using English as a bridge. While large parallel corpora between two languages are scarce, using our approach we can concatenate multiple parallel corpora to obtain a large multilingual corpus. 
The parameters are estimated in a Bayesian nonparametric framework that allows our algorithm to
only associate a word with a new sense vector when evidence (from
either same or foreign language context) requires it. As a result, the
model infers different number of senses for each word in a
data-driven manner, avoiding wasting parameters.

Together, these two ideas -- 
multilingual distributional information and nonparametric sense modeling -- allow us to disambiguate
multiple senses using far less data than is necessary for previous
methods.  We experimentally demonstrate that our algorithm can achieve
competitive performance after training on a small multilingual
corpus, comparable to a model trained monolingually on a much larger
corpus.  We present an analysis discussing the effect of various parameters -- choice
of language family for deriving the multilingual signal,
crosslingual window size etc. and also show qualitative improvement in the embedding
space.



\section{Related Work}
\label{sec:related-work}
Work on inducing multi-sense embeddings can be divided in two broad
categories -- two-staged approaches and joint learning approaches.
Two-staged approaches~\cite{reisinger:naaclx10,huang2012improving} induce multi-sense
embeddings by first clustering the contexts and then using the clustering to obtain the sense vectors. The contexts can be topics induced using latent topic models\cite{liu2015learning,liu2015topical},
or Wikipedia~\cite{wu2015sense} or coarse part-of-speech
tags~\cite{qiu2014learning}.  A more recent line of work in the
two-staged category is that of
retrofitting~\cite{faruqui:2014:NIPS-DLRLW,jauhar2015ontologically},
which aims to infuse semantic ontologies from resources like
WordNet~\cite{miller:1995} and Framenet~\cite{baker1998berkeley} into
embeddings during a post-processing step. Such resources list (albeit
not exhaustively) the senses of a word, and by retro-fitting it is
possible to tease apart the different senses of a word. While some resources like WordNet~\cite{miller:1995} are available for many languages, they are not exhaustive in listing all possible
senses. Indeed, the number senses of a word is highly dependent on the
task and cannot be pre-determined using a
lexicon~\cite{kilgarriff1997don}. Ideally, the senses should be inferred in a data-driven manner, so that new senses not listed in such lexicons can be discovered. 
While recent work has attempted to remedy this by
using parallel text for retrofitting sense-specific
embeddings~\cite{ettinger2016retrofitting}, their procedure requires
creation of {\em sense graphs}, which introduces additional tuning
parameters. On the other hand, our approach only requires two tuning
parameters (prior $\alpha$ and maximum number of senses $T$).

In contrast, joint learning approaches
\cite{neelakantan-EtAl:2014:EMNLP2014,li2015hierarchical} jointly learn the sense clusters and
embeddings by using
non-parametrics. Our approach belongs to this category. The closest
non-parametric approach to ours is that of
\cite{bartunov2015breaking}, who proposed a multi-sense variant of the
skip-gram model which learns the different number of
sense vectors for all words from a large monolingual corpus
(eg. English Wikipedia). Our work can be viewed as the multi-view
extension of their model which leverages both monolingual and
crosslingual distributional signals for learning the embeddings. In
our experiments, we compare our model to monolingually trained
version of their model.

Incorporating crosslingual distributional information is a popular technique
for learning word embeddings, and improves performance on several downstream
tasks~\cite{faruqui-dyer:2014:EACL,guo2016representation,bicompare:16}.
However, there has been little work on learning multi-sense embeddings using crosslingual signals \cite{bansal2012unsupervised,guo2014learning,vsuster-titov-vannoord:2016:N16-1} with only \cite{vsuster-titov-vannoord:2016:N16-1} being a joint approach. \cite{kawakami2015learning} also used bilingual distributional signals in a deep neural architecture to learn context dependent representations for words, though they do not learn separate sense vectors. 


\section{Model Description}
\label{sec:model-description}
Let $E=\{x^e_1,..,x^e_i,..,x^e_{N_e}\}$ denote the words of the English
side and $F=\{x^f_1,..,x^f_i,..,x^f_{N_f}\}$ denote the words of the
foreign side of the parallel corpus. We assume that we have access to word
alignments $A_{e \rightarrow f}$ and $A_{f \rightarrow e}$ mapping
words in English sentence to their translation in foreign sentence
(and vice-versa), so that $x^e$ and $x^f$ are aligned if $A_{e \rightarrow f}(x^e)=x^f$. 

We define Nbr($x,L,d$) as the neighborhood in language $L$ of size $d$ (on either side) around word $x$ in its sentence. The English and foreign neighboring words are denoted by $y^e$ and $y^f$, respectively. Note that $y^e$ and $y^f$ need not be translations of each other.  Each word $x^f$ in the foreign vocabulary is associated with
a dense vector $\bx^f$ in $\mathbb{R}^m$, and each word $x^e$ in
English vocabulary admits at most $T$ sense vectors, with the $k^{th}$
sense vector denoted as $\bx^e_{k}$.\footnote{We
  also maintain a context vector for each word in the English and
  Foreign vocabularies. The context vector is used as the
  representation of the word when it appears as the context for
  another word.}  As our main goal is to model multiple senses for
words in English, we do not model polysemy in the foreign language and
use a single vector to represent each word in the foreign vocabulary.

We model the joint conditional distribution of the context words $y^e, y^f$  given an English word $x^e$ and its corresponding translation $x^f$ on the parallel corpus:
\begin{equation}
\label{eq:cond1}
P(y^e,y^f \mid x^e,x^f;\alpha,\theta),
\end{equation}
where $\theta$ are model parameters (i.e. all embeddings) and $\alpha$ governs the hyper-prior on latent senses.

Assume $x^e$ has multiple senses, which are indexed by the random variable $z$, Eq. \eqref{eq:cond1} can be rewritten, 
\begin{equation*}
\label{eq:latent}
\int_\beta\sum\nolimits_z P(y^e,y^f z,\beta \mid  x^e, x^f, \alpha;\theta)d\beta
\end{equation*}
where $\beta$ are the parameters determining the model probability on each 
sense for $x^e$ (i.e., the weight on each possible value for $z$).  We place a Dirichlet process~\cite{ferguson1973bayesian} prior on sense assignment for each word.  Thus, adding the word-$x$ subscript to emphasize that these are word-specific senses,
\begin{align}\label{eq:dpdef}
  P(z_x=k \mid \beta_x) = \beta_{xk} \prod\nolimits_{r=1}^{k-1} (1- \beta_{xr}) \\
  \beta_{xk} \mid \alpha \stackrel{ind}{\sim} Beta(\beta_{xk} \mid 1, \alpha), ~~k=1,\ldots. 
\end{align}
That is, the potentially infinite number of senses for each word $x$ have probability determined by the sequence of independent {\it stick-breaking weights}, $\beta_{xk}$,  in  the constructive definition of the DP
\cite{sethuraman1994constructive}.  The hyper-prior concentration $\alpha$ provides information on the number of senses we expect to observe in our corpus.

After conditioning upon word sense, we decompose the context probability,
\begin{align*}
  & P(y^e,y^f \mid  z, x^e, x^f; \theta) = \\
  & P(y^e \mid x^e, x^f, z;\theta)  P(y^f \mid x^e, x^f, z;\theta).
\end{align*}

\begin{figure}
  \centering
  \includegraphics[scale=0.3]{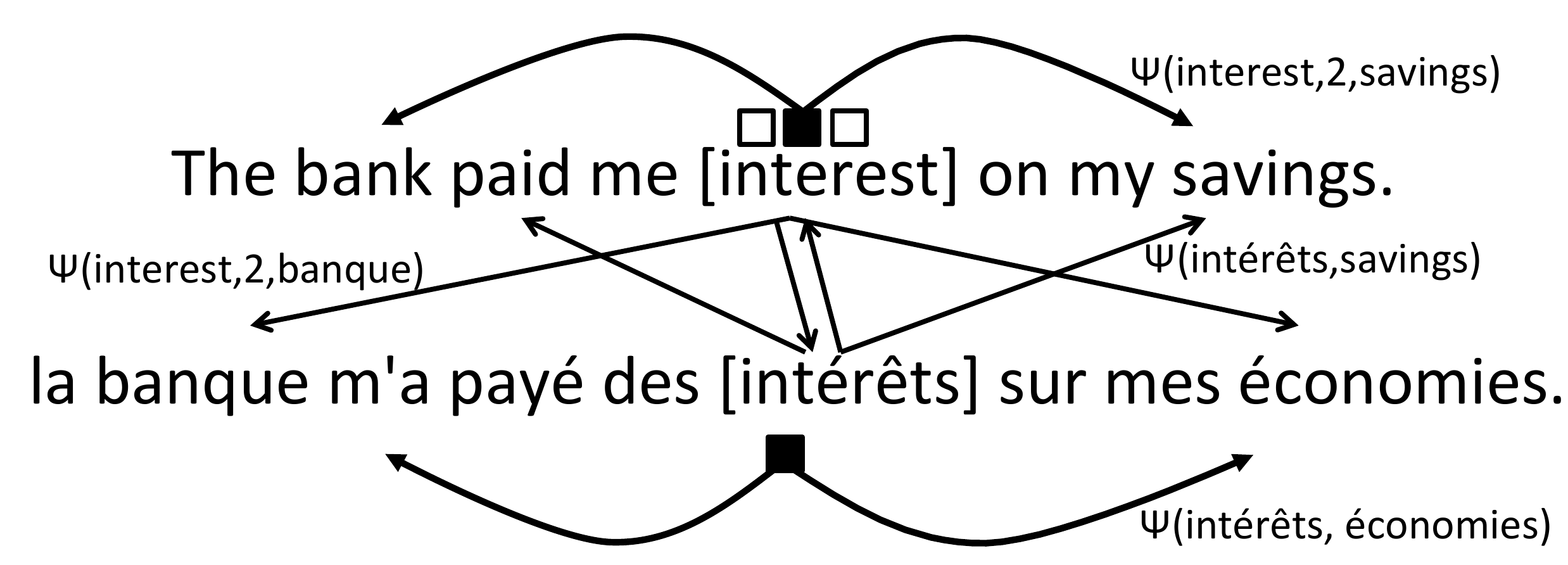}
  \caption{
  The aligned pair ({\em interest},{\em int\'er\^et}) is used to predict monolingual and crosslingual context in both languages (see factors in eqn. \eqref{eq:factors}). We pick each sense (here 2nd) vector for {\em interest}, to perform weighted update. We only model polysemy in English.}
  \label{fig:cartoon}
\end{figure}

Both the first and the second terms are sense-dependent, and each factors as,
\begin{align*}
   P(y\!\mid\! x^e,x^f,z\!=\!k;\theta) \!\propto\! \Psi(x^e,z\!=\!k,y) \Psi(x^f,y) \\
  = \exp({\by^T \bx^e_k})\exp({\by^T \bx^f})\notag = \exp({\by^T(\bx^e_k\!+\!\bx^f)}), \end{align*}
where $\bx^e_k$ is the embedding corresponding to the $k^{th}$ sense
of the word $x^e$, and $y$ is either $y^e$ or $y^f$. The factor
$\Psi(x^e,z=k,y)$ use the corresponding sense vector in a
skip-gram-like formulation. This results in total of 4 factors,
\begin{equation}
\label{eq:factors}
\begin{split}
  P(y^e,y^f \mid  z, x^e, x^f; \theta) & \propto \Psi(x^e,z,y^e) \Psi(x^f,y^f) \\
  & \Psi(x^e,z,y^f) \Psi(x^f,y^e)
\end{split}
\end{equation}
See Figure \ref{fig:cartoon} for illustration of each factor.
This modeling approach is reminiscent of~\cite{luong:2015}, who
jointly learned embeddings for two languages $l_1$ and $l_2$ by
optimizing a joint objective containing 4 skip-gram terms using the
aligned pair ($x^e$,$x^f$)-- two predicting monolingual contexts $l_1 \rightarrow
l_1$, $l_2 \rightarrow l_2$ , and two predicting crosslingual
contexts $l_1 \rightarrow l_2$, $l_2 \rightarrow l_1$. 

\paragraph{Learning.}
Learning involves maximizing the log-likelihood,
\begin{align*}
  & P(y^e,y^f \mid x^e,x^f;\alpha,\theta) =\\
  & \int_\beta\sum\nolimits_z P(y^e,y^f,z,\beta \mid x^e, x^f, \alpha;\theta)d\beta
\end{align*}
for which we use variational approximation.
Let $q(z,\beta)=q(z)q(\beta)$ where
\begin{equation}
\label{equ:full-proj}
\begin{array}{cc}
q(z)=\prod_i q(z_i) & q(\beta)=\prod_{w=1}^V\prod_{k=1}^T \beta_{wk}
\end{array}
\end{equation}
are the fully factorized variational approximation of the true
posterior $P(z,\beta \mid y^e,y^f,x^e,x^f,\alpha)$, where $V$ is the
size of english vocabulary, and $T$ is the maximum number of senses
for any word. The optimization problem solves for $\theta$,$q(z)$ and
$q(\beta)$ using the stochastic variational inference
technique~\cite{hoffman2013stochastic} similar to
\cite{bartunov2015breaking} (refer for details).

The resulting learning algorithm is shown as Algorithm
\ref{algo:model}. The first for-loop (line \ref{lst:line:es}) updates
the English sense vectors using the crosslingual and monolingual
contexts. First, the expected sense distribution for the current
English word $w$ is computed using the current estimate of $q(\beta)$
(line \ref{lst:line:estep}).  The sense distribution is updated (line
\ref{lst:line:sense}) using the combined monolingual and
crosslingual contexts (line \ref{lst:line:combine}) and re-normalized
(line \ref{lst:line:renorm}). Using the updated sense distribution
$q(\beta)$'s sufficient statistics is re-computed (line
\ref{lst:line:suff}) and the global parameter $\theta$ is updated
(line \ref{lst:line:theta}) as follows,
\begin{equation}
   \label{eq:thetaup}
  \theta \leftarrow \theta + \rho_t \nabla_{\theta}\sum\limits_{k \mid z_{ik} > \epsilon} \sum\limits_{y \in y_c} z_{ik} \log
  p(y|x_i,k,\theta)
\end{equation}
Note that in the above sum, a sense participates in a update only if
its probability exceeds a threshold $\epsilon$ (= 0.001). The final
model retains sense vectors whose sense probability exceeds the
same threshold.
The last for-loop (line \ref{lst:line:fs}) jointly
optimizes the foreign embeddings using English context with the standard skip-gram updates.


\begin{algorithm}[t]
\caption{Psuedocode of Learning Algorithm}
\begin{algorithmic}[1]
\Require{parallel corpus $E=\{x^e_1,..,x^e_i,..,x^e_{N_e}\}$ and $F=\{x^f_1,..,x^f_i,..,x^f_{N_f}\}$ and alignments $A_{e \rightarrow f}$ and $A_{f \rightarrow e}$, Hyper-parameters $\alpha$ and $T$, window sizes $d,d'$
}.
\Ensure{$\theta$, $q(\beta)$, $q(\mathbf{z})$}
\For{$i=1$ to $N_e$}\label{lst:line:es}
\Comment{update english vectors}
\Let{$w$}{$x^e_{i}$}
\For{$k=1$ to $T$}
\Let{$z_{ik}$}{$\mathbb{E}_{q(\beta_w)}[\log p(z_i = k| ,x^e_{i})]$} \label{lst:line:estep}
\EndFor
\Let{$y_c$}{Nbr($x^e_i$,$E$,$d$) $\cup$ Nbr($x^f_i$,$F$,$d'$) $\cup$ \{$x^f_i$\} where $x^f_i=A_{e\rightarrow f}(x^e_i)$} \label{lst:line:combine}
\For{$y$ in $y_c$}
\State{\Call{sense-update}{$x^e_i,y,z_i$}} \label{lst:line:sense}
\EndFor
\State{Renormalize $z_i$ using softmax} \label{lst:line:renorm}
\State{Update suff. stats. for $q(\beta)$ like \cite{bartunov2015breaking}} \label{lst:line:suff}
\State{Update $\theta$ using eq. \eqref{eq:thetaup}} \label{lst:line:theta}
\EndFor
\For{$i=1$ to $N_f$}\label{lst:line:fs}
\Comment{jointly update foreign vectors}
\Let{$y_c$}{Nbr($x^f_i$,$F$,$d$) $\cup$ Nbr($x^e_i$,$E$,$d'$) $\cup$ \{$x^e_i$\} where $x^e_i=A_{f\rightarrow e}(x^f_i)$}\label{lst:line:value}
\For{y in $y_c$}
\State{{\sc skip-gram-update}{$(x^f_i,y)$}}
\EndFor
\EndFor

\Procedure{sense-update}{$x_i,y,z_i$}
\Let{$z_{ik}$}{$z_{ik} + \log p(y| x_i, k, \theta)$}
\EndProcedure
\end{algorithmic}
\label{algo:model}
\end{algorithm}

\paragraph{Disambiguation.}
Similar to \cite{bartunov2015breaking}, we can disambiguate the sense
for the word $x^e$ given a monolingual context $y^e$ as follows,
\begin{equation}
\begin{split}
  &P(z\mid x^e, y^e)\propto  \\
  &P(y^e \mid x^e, z;\theta)\sum\nolimits_{\beta}
  P(z \mid x^e, \beta) q(\beta) \label{eq:disamb}
\end{split}
\end{equation}
Although the model trains embeddings using both monolingual
and crosslingual context, we only use monolingual context at test time. We found that 
so long as the model has been trained with multilingual context, it performs well in sense disambiguation on new data even if it contains only monolingual context. A similar observation was made
by~\cite{vsuster-titov-vannoord:2016:N16-1}.


\section{Multilingual Extension}
\label{sec:multi-ext}
Bilingual distributional signal
alone may not be sufficient as polysemy may survive translation in the second language. Unlike existing approaches, we can easily incorporate multilingual
distributional signals in our model. For using languages $l_1$ and $l_2$ to
learn multi-sense embeddings for English, we train on a
concatenation of En-$l_1$ parallel corpus with an En-$l_2$ parallel
corpus. This technique can easily be generalized to more than two foreign
languages to obtain a large multilingual corpus.

\paragraph{Value of $\Psi(y^e, x^f)$.}
The factor modeling the dependence of the English context word $y^e$
on foreign word $x^f$ is crucial to performance when using multiple
languages. Consider the case of using French and Spanish
contexts to disambiguate the financial sense of the English word {\em
  bank}. In this case, the (financial) sense vector of {\em bank} will
be used to predict vector of {\em banco} (Spanish context) and
{\em banque} (French context). If vectors for {\em banco} and
{\em banque} do not reside in the same space or are not close, the model will incorrectly assume they are different contexts
to introduce a new sense for 
{\em bank}. This is precisely why the bilingual models, like that of \cite{vsuster-titov-vannoord:2016:N16-1}, cannot be extended to multilingual setting, as they pre-train the embeddings of second language before running the multi-sense embedding process. As a result of naive pre-training, the French and Spanish vectors of semantically similar pairs like ({\em banco},{\em banque}) will lie in different spaces and need not be close. 
A similar reason holds for \cite{guo2014learning}, as they use a two step approach instead of joint learning. 

To avoid this, the vector for pairs like {\em
  banco} and {\em banque} should lie in the same space and close to
each other and the sense vector for {\em bank}. The $\Psi(y^e, x^f)$
term attempts to ensure this by using the vector for {\em banco} and
{\em banque} to predict the vector of {\em bank}. This way, the model
brings the embedding space for Spanish and French closer by using English as a
bridge language during joint training. A similar idea of using English as a bridging language
was used in the models proposed
in~\cite{Hermann:2014:ICLR} and \cite{coulmance-EtAl:2015:EMNLP}. Beside the benefit in the multilingual
case, the $\Psi(y^e, x^f)$ term improves performance in the
bilingual case as well, as it forces the English and second language
embeddings to remain close in space.

To show the value of $\Psi(y^e,x^f)$ factor in our experiments,
we ran a variant of Algorithm~\ref{algo:model} without the
$\Psi(y^e,x^f)$ factor, by only using monolingual
neighborhood $Nbr(x^f_i,F)$ in line \ref{lst:line:value} of
Algorithm~\ref{algo:model}. We call this variant {\sc One-Sided} model and the model in Algorithm~\ref{algo:model} the {\sc Full} model. 


\section{Experimental Setup}
\label{sec:experimental-setup}
\begin{table}
\centering
  \begin{tabular}{l@{ }l@{ }l@{ }c}
    Corpus & Source & Lines (M) & EN-Words (M)\\
    \toprule
    En-Fr &  EU proc. & $\approx 10$ & 250\\
    \midrule
    En-Zh & FBIS news &  $\approx 9.5$ & 286\\
    \midrule
    En-Es &  UN proc. & $\approx 10$ & 270\\
    En-Fr &  UN proc. & $\approx 10$ & 260\\
    En-Zh &  UN proc. & $\approx 8$ & 230\\
    En-Ru &  UN proc. & $\approx 10$ & 270\\
    \bottomrule
  \end{tabular}
  \caption{Corpus Statistics (in millions). Horizontal lines demarcate
    corpora from the same domain.}
    \label{tab:stats}
\end{table}
We first describe the datasets and the preprocessing
methods used to prepare them. We also describe the Word Sense Induction task that we used to compare and evaluate our method. 

\paragraph{Parallel Corpora.}
We use parallel corpora in English (En), French (Fr), Spanish (Es), Russian (Ru) and Chinese (Zh) in our experiments. 
Corpus statistics for all datasets used in our experiments
are shown in Table \ref{tab:stats}.
For En-Zh, we use the FBIS parallel corpus (LDC2003E14). For En-Fr, we use the first 10M lines
from the Giga-EnFr corpus released as part of the WMT shared
task~\cite{callison2011findings}.
Note that the domain from which parallel corpus has been derived can
affect the final result. To understand what choice of languages
provide suitable disambiguation signal, it is necessary to control for
domain in all parallel corpora. To this end, we also used the En-Fr,
En-Es, En-Zh and En-Ru sections of the MultiUN parallel
corpus~\cite{multiUN}. Word alignments 
were generated using \texttt{fast_align}
tool~\cite{dyer2013simple} in the symmetric intersection
mode. Tokenization and other preprocessing were performed using {\tt
  cdec}~\footnote{\url{github.com/redpony/cdec}} toolkit. Stanford Segmenter~\cite{I05-3027} was used to preprocess the Chinese corpora.

\paragraph{Word Sense Induction (WSI).}
We evaluate our approach on word sense induction task. In this task,
we are given several sentences showing usages of the same word, and
are required to cluster all sentences which use the same
sense~\cite{nasiruddin2013state}. The predicted clustering is then
compared against a provided gold clustering. Note that WSI is a harder
task than Word Sense Disambiguation (WSD)\cite{navigli2009word}, as
unlike WSD, this task does not involve any supervision or explicit
human knowledge about senses of words. We use the disambiguation
approach in eq.~(\ref{eq:disamb}) to predict the sense given the
target word and four context words.

To allow for fair comparison with earlier work, we use the same
benchmark datasets as~\cite{bartunov2015breaking} -- Semeval-2007,
2010 and Wikipedia Word Sense Induction (WWSI). We report
Adjusted Rand Index  (ARI)~\cite{hubert1985comparing}
in the experiments, as ARI is a more strict
and precise metric than F-score and V-measure.

\paragraph{Parameter Tuning.}
For fairness, we used five context words on either side to update each
English word-vectors in all the experiments. In the monolingual setting, all five words are
English; in the multilingual settings, we used four neighboring
English words plus the one foreign word aligned to the word being
updated ($d=4$, $d'=0$ in Algorithm~\ref{algo:model}). We also analyze effect of varying $d'$, the context window size in the foreign sentence on the model performance. 

We tune the parameters $\alpha$ and $T$ by maximizing the log-likelihood of a held out
English text.\footnote{first 100k lines
from the En-Fr Europarl~\cite{koehn2005europarl}} The parameters were chosen from the following values
$\alpha=\{0.05,0.1,..,0.25\}$,
$T=\{5,10,..,30\}$. All models were trained for 10 iteration
with a decaying learning rate of 0.025, decayed to 0. Unless otherwise stated, all embeddings are 100
dimensional.

Under various choice of $\alpha$ and $T$, we identify only about 10-20\% polysemous words in the vocabulary using monolingual training and 20-25\% polysemous using multilingual training. It is evident using the non-parametric prior has led to substantially more efficient representation compared to previous methods with fixed number of senses per word.   



\section{Experimental Results}
\label{sec:experimental-results}
\begin{table}[ht]
  \centering
  \footnotesize
    \begin{tabular}{@{}l@{ }c@{\quad}c@{\quad}c@{\quad}c|c@{}}
      Setting & S-2007 & S-2010 & WWSI & avg. ARI & SCWS\\
      \toprule
      \multicolumn{6}{c}{En-Fr} \\
      \midrule
      {\sc Mono} &  .044 & .064 & .112 & .073 & 41.1\\
      {\sc One-Sided} & .054 & .074 & {\bf .116} & .081 & {\bf 41.9}\\
      {\sc Full} &  {\bf .055} & {\bf .086} & .105 & {\bf .082} & 41.8\\
      \midrule
      \multicolumn{6}{c}{En-Zh} \\
      \midrule
      {\sc Mono} & .054 & .074 & .073 & .067 & 42.6\\
      {\sc One-Sided} & {\bf .059} & .084 & .078 & .074 & {\bf 45.0}\\
      {\sc Full} & .055 & {\bf .090} & {\bf .079} & {\bf .075} & 41.7\\
      \midrule
      \multicolumn{6}{c}{En-FrZh} \\
      \midrule
      {\sc Mono} &  .056 & .086 & .103 & .082 & {\bf 47.3}\\
      {\sc One-Sided} & {\bf .067} & .085 & .113 & .088 & 44.6\\
      {\sc Full} &  .065 & {\bf .094} & {\bf .120} & {\bf .093} & 41.9\\
      \bottomrule
    \end{tabular}
    \caption{Results on word sense induction (left four columns) in
      ARI and contextual word similarity (last column) in percent
      correlation. Language pairs are separated by horizontal lines. Best results  shown in {\bf bold}.}
    \label{tab:wsi}
\end{table}
\begin{table*}
  \centering
  \footnotesize
  \begin{tabular}{lllllllll}
  \toprule
    Train & \multicolumn{2}{c}{S-2007} & \multicolumn{2}{c}{S-2010} & \multicolumn{2}{c}{WWSI} & \multicolumn{2}{c}{Avg. ARI}\\
    Setting & En-FrEs & En-RuZh & En-FrEs & En-FrEs & En-FrEs & En-RuZh & En-FrEs & En-RuZh\\
    \toprule
    (1) {\sc Mono}  & .035 & .033 & .046 & .049 & .054 & .049 & .045 & .044\\
    (2) {\sc One-Sided} & .044 & {\bf .044} & .055 & .063 & .062 & .057 & .054 & .055\\
    (3) {\sc Full}  & {\bf .046} & .040 & {\bf .056} & {\bf .070} & {\bf .068} & {\bf .069} & {\bf .057} & {\bf .059}\\
    \midrule
    (3) - (1) & .011 & .007 & .010 & .021 & .014 & .020 & .012 & .015\\
    \bottomrule
  \end{tabular}
  \caption{Effect (in ARI) of language family distance on WSI task. Best results for each column is shown in {\bf bold}. The improvement from {\sc Mono} to {\sc Full} is also shown as (3) - (1). Note that this is not comparable to results in Table \ref{tab:wsi}, as we use a different training corpus to control for the domain.}
  \label{tab:family}
\end{table*}

We performed extensive experiments to evaluate the benefit of leveraging bilingual
and multilingual information during training. We also analyze how  the different choices of language family (i.e. using more distant vs more similar languages) affect  performance of the embeddings. 


\subsection{Word Sense Induction Results.}
The results for WSI are shown in Table~\ref{tab:wsi}. Recall that the
{\sc One-Sided} model is the variant of Algorithm~\ref{algo:model}
without the $\Psi(y^e,x^f)$ factor. {\sc Mono} refers to the AdaGram
model of~\cite{bartunov2015breaking} trained on the English side of
the parallel corpus.  In all cases, the {\sc Mono} model is
outperformed by {\sc One-Sided} and {\sc Full} models, showing the
benefit of using crosslingual signal in training. Best performance is
attained by the multilingual model (En-FrZh), showing value of
multilingual signal. The value of $\Psi(y^e,x^f)$ term is also
verified by the fact that the {\sc One-Sided} model performs worse
than the {\sc Full} model.

We can also compare (unfairly to our {\sc Full} model) to the best
results described in~\cite{bartunov2015breaking}, which achieved ARI
scores of 0.069, 0.097 and 0.286 on the three datasets respectively
after training 300 dimensional embeddings on English Wikipedia
($\approx$ 100M lines). Note that, as WWSI was derived from Wikipedia,
training on Wikipedia gives AdaGram model an undue advantage,
resulting in high ARI score on WWSI. In comparison, our model did not train on English Wikipedia, and uses 100 dimensional embeddings. Nevertheless, even in the unfair
comparison, it noteworthy that on S-2007 and S-2010, we can achieve
comparable performance (0.067 and 0.094) with multilingual training to
a model trained on almost 5 times more data using higher (300)
dimensional embeddings.

\subsection{Contextual Word Similarity Results.} 
For completeness, we report correlation scores on Stanford contextual word similarity dataset (SCWS)~\cite{huang2012improving} in Table~\ref{tab:wsi}. The task requires computing similarity between two words given their contexts. While the bilingually trained model outperforms the monolingually trained model, surprisingly the multilingually trained model does not perform well on SCWS. We believe this may be due to our parameter tuning strategy.\footnote{Most works tune directly on the test dataset for Word Similarity tasks~\cite{repeval:16}}

\subsection{Effect of Language Family Distance.}
\begin{figure*}[ht]
  \centering
    \begin{subfigure}[b]{0.45\linewidth}
    \centering
    \includegraphics[scale=0.42]{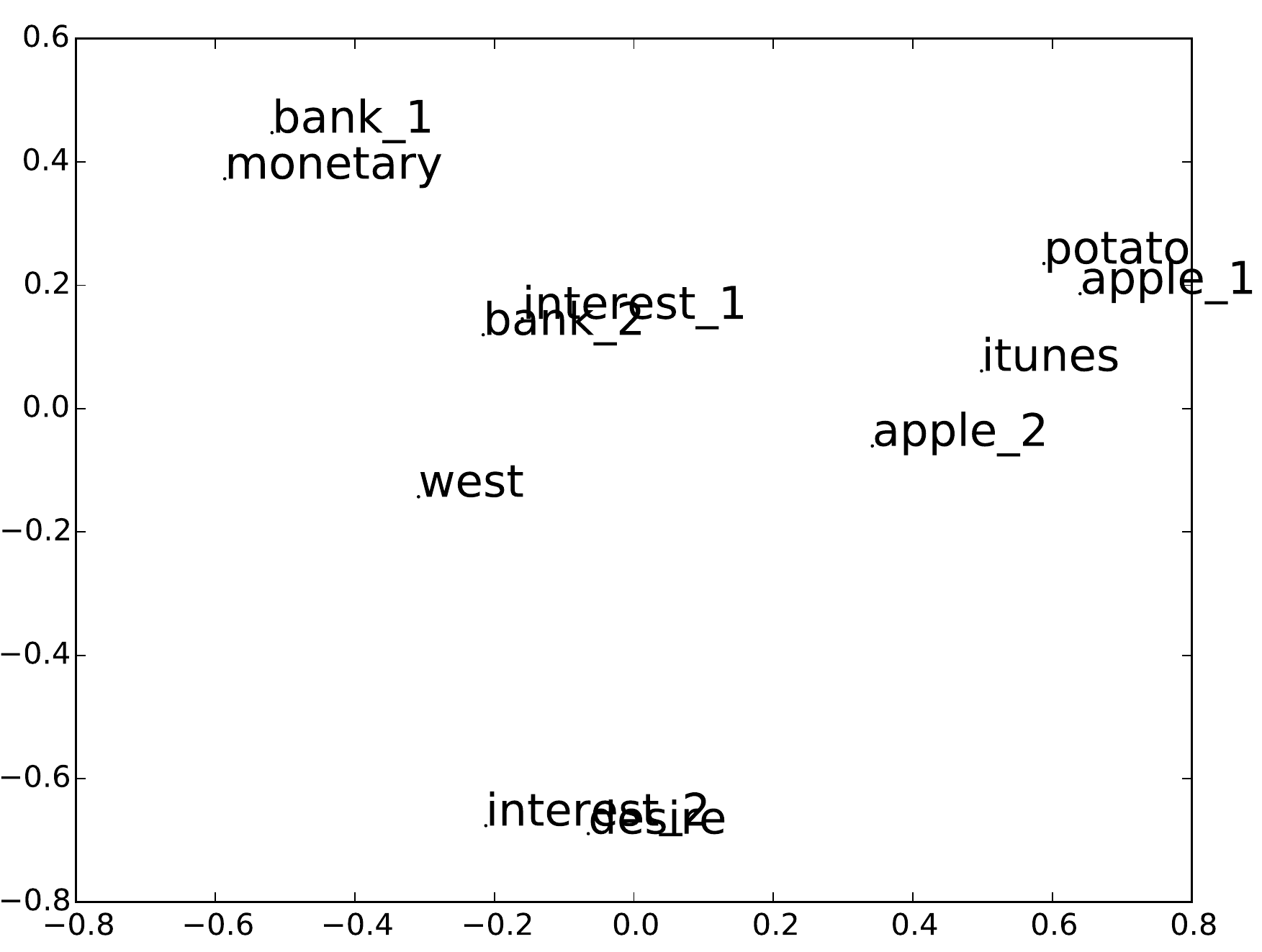}
    \caption{Monolingual (En side of En-Zh)}
    \label{fig:mono}
  \end{subfigure}
   \hfill
  \begin{subfigure}[b]{0.45\linewidth}
    \centering
    \includegraphics[scale=0.42]{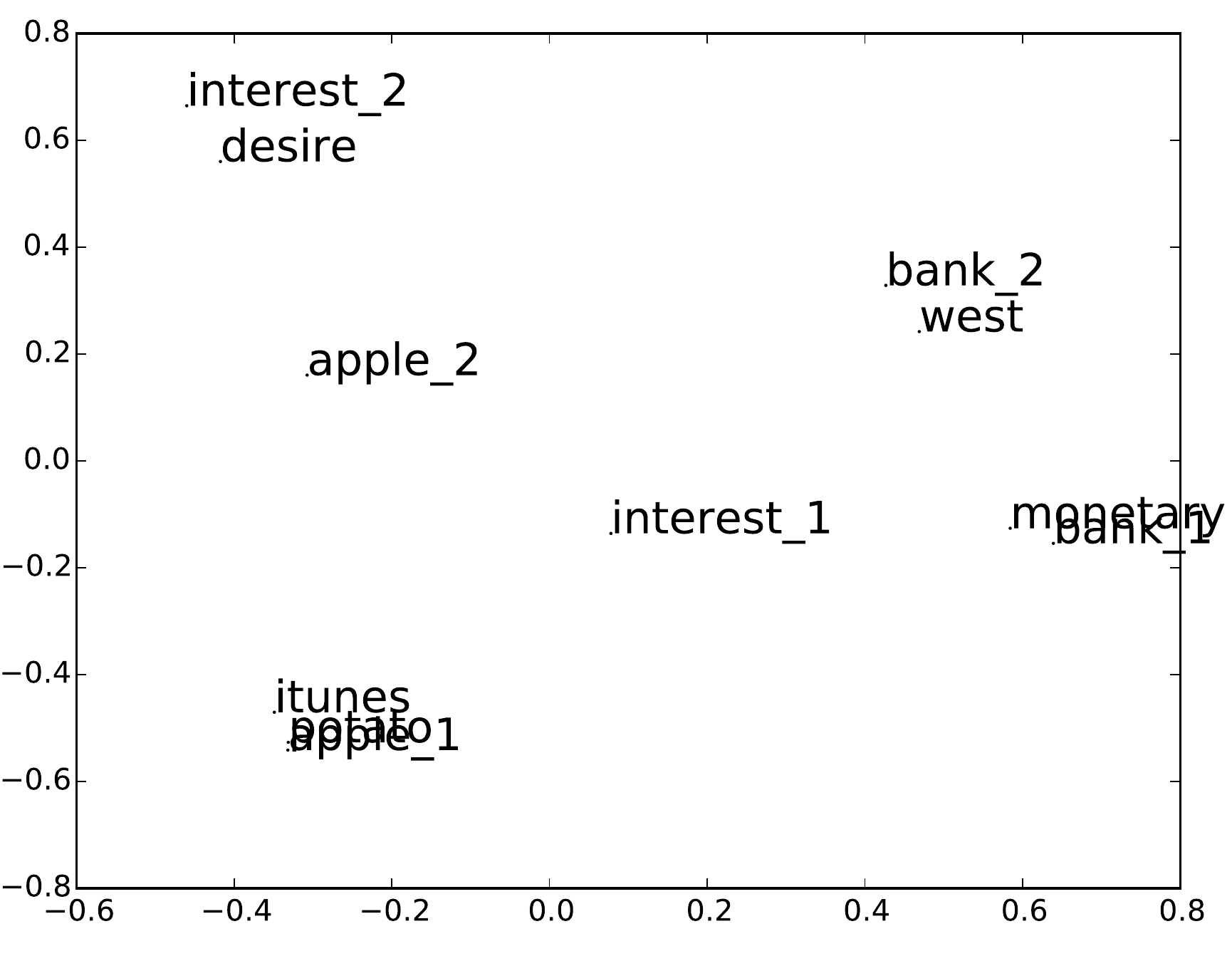}
    \caption{Bilingual (En-Zh)}
    \label{fig:bi}
  \end{subfigure}
  \vskip\baselineskip
  \begin{subfigure}[b]{0.45\linewidth}
      \centering
    \includegraphics[scale=0.42]{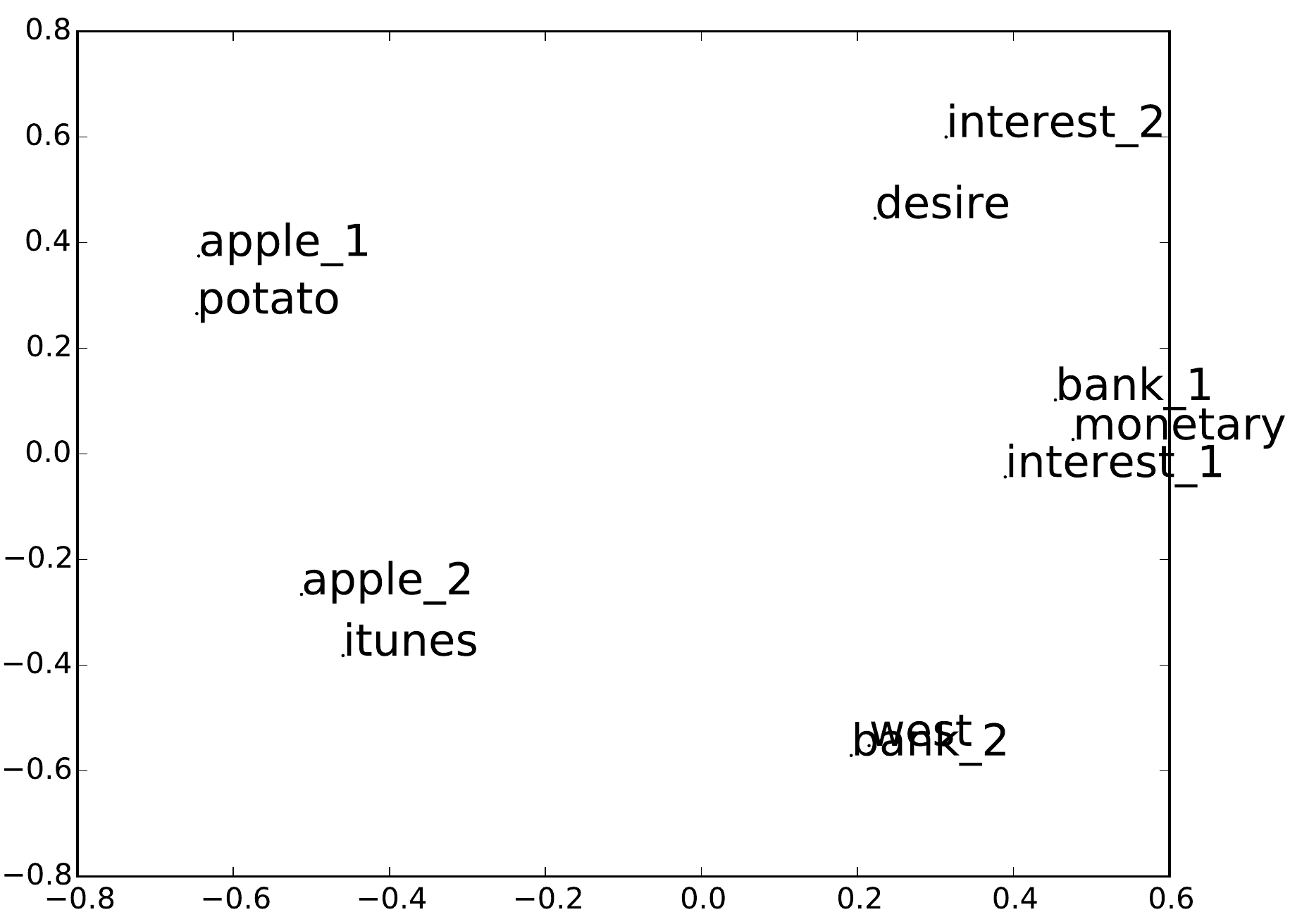}
    \caption{Multilingual (En-FrZh)}
    \label{fig:tri}
  \end{subfigure}
  \hfill
  \begin{subfigure}[b]{0.45\linewidth}
  \centering
  \resizebox{6.5cm}{6cm}{
    \begin{tikzpicture} 
      \begin{axis}[
          xlabel=window,
      ylabel=avg. ARI,
      legend style={
        at={(0.8,0.3)},       
        anchor=north,
        legend columns=1       
      },
    ]
        \addplot table [x=a, y=b, col sep=comma] {tune.csv};
        \addplot table [x=a, y=c, col sep=comma] {tune.csv};
        \legend{En-Fr,En-Zh}
      \end{axis}
\end{tikzpicture}
  }
  \caption{Window size v.s. avg. ARI }
  \label{fig:wtune}
\end{subfigure}

  \caption{{\bf Qualitative:} PCA plots for the vectors for {\{\em apple, bank, interest, itunes, potato, west, monetary, desire\}} with multiple sense vectors for {\em apple},{\em interest} and {\em bank} obtained using monolingual (\ref{fig:mono}), bilingual (\ref{fig:bi}) and multilingual (\ref{fig:tri}) training. {\bf Window Tuning:} Figure \ref{fig:wtune} shows tuning window size for En-Zh and En-Fr.}
    \label{fig:qual}
\end{figure*}
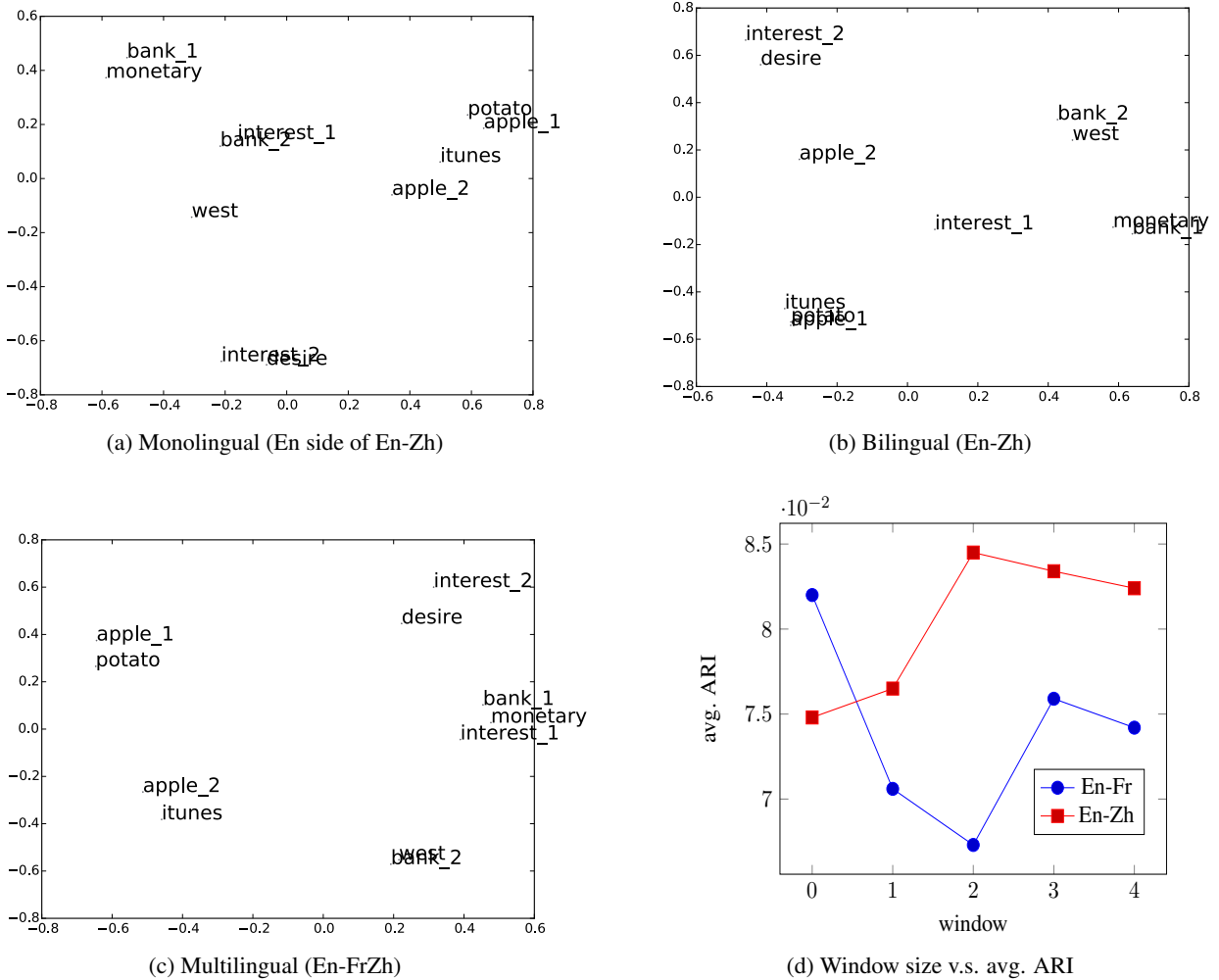

Intuitively, choice of language can affect the result from
crosslingual training as some languages may provide better disambiguation signals
than others.  We performed a systematic set of experiment to evaluate whether we
should choose languages from a closer family (Indo-European
languages) or farther family (Non-Indo European Languages) as training data alongside English.\footnote{~\cite{vsuster-titov-vannoord:2016:N16-1}
  compared different languages but  did not control for
  domain.}
To control for domain here we use the MultiUN corpus. We use En
paired with Fr and Es as Indo-European languages, and English paired
with Ru and Zh for representing Non-Indo-European languages.

From Table~\ref{tab:family}, we see that using Non-Indo European
languages yield a slightly higher improvement on an average than using Indo-European languages. This suggests that using
languages from a distance family aids better disambiguation. Our
findings echo those of \cite{resnik1999distinguishing}, who found that
the tendency to lexicalize senses of an English word differently in a
second language, correlated with language
distance. 

\subsection{Effect of Window Size.}
Figure~\ref{fig:wtune} shows the effect of increasing the
crosslingual window ($d'$) on the average ARI on the WSI task for
the En-Fr and En-Zh models.
While increasing the window
size improves the average score for En-Zh model, the score for the
En-Fr model goes down. 
This suggests that it might be beneficial to have a separate window
parameter per language. This also aligns with the observation earlier
that different language families have different suitability (bigger
crosslingual context from a distant family helped) and requirements
for optimal performance.

 

\section{Qualitative Illustration}
\label{sec:qualitative-analysis}
As an illustration for the effects of multilingual training, Figure \ref{fig:qual} shows PCA plots for 11 sense vectors for 9
words 
using monolingual, bilingual and multilingual models.
From Fig~\ref{fig:mono}, we note that with
monolingual training the senses are poorly separated. Although the model infers two senses for {\em bank},
the two senses of {\em bank} are close to financial terms, suggesting their distinction was not recognized. The same observation can be made for the senses of {\em apple}.
In Fig~\ref{fig:bi}, with bilingual training, the model infers two
senses of {\em bank} correctly, and two sense of {\em apple} become more
distant. The model can still improve eg. pulling
{\em interest} towards the financial sense of {\em bank}, and pulling {\em itunes}
towards {\em apple\_2}.
Finally, in Fig~\ref{fig:tri}, all senses of the words are more
clearly clustered, improving over the clustering of Fig
\ref{fig:bi}. The senses of {\em apple}, {\em interest}, and {\em bank} are well
separated, and are close to sense-specific words, showing the benefit of multilingual training.

\section{Conclusion}
\label{sec:conclusion}
We presented a multi-view, non-parametric word representation learning
algorithm which can leverage multilingual distributional
information. Our approach effectively combines the benefits of
crosslingual training and Bayesian non-parametrics. Ours is the first
multi-sense representation learning algorithm capable of using
multilingual distributional information efficiently, by combining
several parallel corpora to obtained a large multilingual corpus. Our
experiments show how this multi-view approach learns high-quality
embeddings using substantially less data and parameters than prior
state-of-the-art.  We also analyzed the effect of various parameters
such as choice of language family and cross-lingual window size on the
performance.  While we focused on improving the embedding of English
words in this work, the same algorithm could learn better multi-sense
embedding for other languages. Exciting avenues for future research
include extending our approach to model polysemy in foreign language.
The sense vectors can then be aligned across languages, to generate a
multilingual Wordnet like resource, in a completely unsupervised
manner thanks to our joint training paradigm.

\bibliography{ref}
\bibliographystyle{acl_natbib}

\end{document}